\documentclass[10pt,twocolumn,letterpaper]{article} 

\usepackage{avss}
\usepackage{times}
\usepackage{epsfig}
\usepackage{graphicx}
\usepackage{amsmath}
\usepackage{amssymb}

\usepackage{enumitem}
\usepackage{tablefootnote}
\usepackage{multirow}
\usepackage{multicol}
\usepackage{capt-of}
\usepackage{fixltx2e}
\usepackage{floatrow}

\usepackage{pifont}
\newcommand{\sub}{\textsubscript} 
\newcommand{\UCF}{UCF\textunderscore CROWD\textunderscore 50}

\usepackage[pagebackref=true,breaklinks=true,letterpaper=true,colorlinks,bookmarks=false]{hyperref}

\avssfinalcopy 

\def\avssPaperID{****} 

\ifavssfinal\fi
\begin{document}

\title{Inverse Attention Guided Deep Crowd Counting Network}

\author{Vishwanath A. Sindagi \qquad Vishal M. Patel\\
	Department of Electrical and Computer Engineering,\\
	Johns Hopkins University, 3400 N. Charles St, Baltimore, MD 21218, USA\\
	{\tt\small \{vishwanathsindagi,vpatel36\}@jhu.edu}
}
\maketitle

\begin{abstract}
In this paper, we address the challenging problem of crowd counting in congested scenes. Specifically, we present  Inverse Attention Guided Deep Crowd Counting Network (IA-DCCN) that  efficiently infuses segmentation information through an inverse attention mechanism into the counting network, resulting in significant improvements. The proposed method, which is based on VGG-16, is a single-step training framework and is  simple to implement.  The use of segmentation information results in minimal computational overhead and does not require any additional annotations. We demonstrate the significance of  segmentation guided   inverse attention through a detailed analysis and ablation study.   Furthermore, the proposed method is  evaluated on three challenging crowd counting datasets and is  shown to achieve  significant improvements over  several recent methods. 

\end{abstract}

\section{Introduction}

Crowd counting \cite{li2015crowded,zhan2008crowd,idrees2013multi,zhang2015cross,zhang2016single,sindagi2017generating,sam2017switching,chan2008privacy,rodriguez2011density,zhu2014crowd,li2014anomaly,mahadevan2010anomaly, marsden2018people, sindagi2019dafe,sam2018top} has attracted a lot of interest in the recent years. With growing population and occurrence of numerous crowded events such as political rallies, protests, marathons, \etc, computer vision-based crowd analysis is becoming an increasingly important task.

Crowd counting suffers from several  challenges such as  scale changes, heavy occlusion, illumination changes, clutter, non-uniform distribution of people, etc.,  making crowd counting and density estimation a very challenging problem, especially in highly congested scenes. Different techniques have been developed to address these issues. The issue of scale variations has received the most interest, with several works proposing different approaches such as multi-column networks \cite{zhang2016single}, switching-cnns \cite{sam2017switching}, use of context information \cite{sindagi2017generating}, \etc. While these methods provide significant improvements over recent techniques, the  error rates of most of these methods are still far from optimal \cite{onoro2016towards,zhang2016single}. A probable reason is that most of these methods train their networks from scratch and since the datasets have limited samples, they are unable to use high-capacity networks.  For the few methods \cite{sindagi2017generating,sam2017switching} that achieve very low error rate, the training process is increasingly complex and requires multiple stages. For instance, Switching-CNN \cite{sam2017switching} involves different stages such as pre-training, differential training, switch training and coupled training. Similarly, CP-CNN \cite{sindagi2017generating} requires that their local and global estimators to be trained separately, followed by end-to-end training of their density map estimator. Although these methods achieve low error, their complex training process makes them hard to use. 

\begin{figure}[t]	
	\centering	
	\includegraphics[width=0.32\linewidth]{figures/fafter/inoutput_31}
	\includegraphics[width=0.32\linewidth]{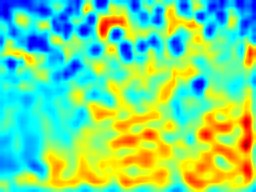}
	\includegraphics[width=0.32\linewidth]{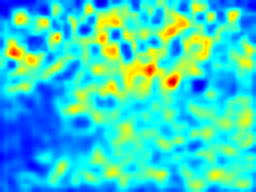}
	(a) \hskip60pt(b) \hskip60pt(c) 
	\vskip-5pt
	\caption{Feature map visualization: (a) Input image, (b) Feature map before refinement,  (c)  Feature map after refinement using inverse attention. By infusing segmentation information via inverse attention into the counting network, we are able to suppress background regions, thus making the counting task much easier.}
	\label{fig:fmap}	
\end{figure}

Considering these drawbacks, our aim in the paper is to design a simple solution that is easy to train and achieves low count error. Given this objective, we start by presenting a VGG-16 based crowd counting network, which alone is able to achieve results that are comparable to recent state-of-the-art methods. While this baseline network achieves comparable performance with respect to recent methods, there is considerable room for further improvement. We present a simple, yet powerful technique that uses multi-task learning to further boost the counting performance. Specifically,  we aim to efficiently infuse foreground/background segmentation mask into the counting network by simultaneously learning to count and segment. This use of related tasks for improving the performance is inspired by the success of recent multi-task approaches such as Hyperface \cite{ranjan2016hyperface}, instance aware semantic segmentation \cite{dai2016instance} and use of semantic segmentation for improving object detection. 

Although this  approach of infusing segmentation information through simple multi-task learning achieves considerable improvements in performance, it is limited by the fact that VGG16 is pre-trained on image net dataset and it will concentrate on regions with high response values during  learning.  To address this issue, we draw inspiration from  the success of attention learning in various tasks such as action recognition, object recognition, image captioning, visual question answering \etc \cite{chen2017sca,song2018mask,sharma2015action,ba2014multiple,liu2017end,you2016image,lu2016hierarchical,xiao2015application,chen2018reverse}, \etc. Specifically, we propose an inverse attention module that captures important regions in the feature maps to focus on during learning.  The inverted attention map enforces the network to focus specifically on relevant regions, thereby increasing the effectiveness of the  learning mechanism.

Through a detailed  ablation study, we demonstrate that the infusion of segmentation information via inverse attention results in enrichment of feature maps there by providing considerable gains in the count error. Fig. \ref{fig:fmap} visualizes the feature maps from intermediate layer  of the base network before and after segmentation infusion via inverse attention. It can be easily observed that by incorporating segmentation information, we are able to suppress the background regions easily. More visualization results are provided in the results section. Furthermore, since the infusion  requires minimal  additional parameters, hence it resulting  in minimal computational overhead during inference.

To summarize, the following are our key contributions:
\begin{itemize}[topsep=0pt,noitemsep,leftmargin=*]
	\item We propose a  crowd counting network that efficiently infuses segmentation information into the counting network.  
	\item  An inverse attention mechanism is introduced to further improve the efficacy of  the learning mechanism.  
\end{itemize}

\vspace{0.2cm}
In the following sections, we discuss related work (Section \ref{sec:related}) and the proposed method in detail (Section \ref{sec:method}). Details of experiments and results of the proposed method along with comparison on different datasets are provided in Section \ref{sec:experiments}, followed by conclusions in Section \ref{sec:conclusion}.

\section{Related work}
\label{sec:related}

\noindent\textbf{Crowd Counting.} Some of the early methods for crowd counting were based on detection techniques \cite{li2008estimating,loy2013crowd}. However, these methods were not robust to occlusions in crowded scenes. To overcome this, several works \cite{ryan2009crowd,chen2012feature,idrees2013multi}  extracted hand designed features from image patches and employed them in different regression frameworks. Since these approaches mapped image patches to count directly, they tend to loose spatial information in the images. This was overcome by the density estimation techniques \cite{lempitsky2010learning,pham2015count,xu2016crowd}, where the counting problem is posed as a pixel-to-pixel translation. A more comprehensive survey of different crowd counting methods can be found in \cite{chen2012feature,li2015crowded}. These methods relied on hand-designed features and hence have limited abilities to achieve low count error. 

Advancements in deep learning and convolutional neural networks (CNN) have boosted  the accuracy of counting techniques. For example several works like  \cite{wang2015deep,zhang2015cross,sam2017switching,arteta2016counting,walach2016learning,onoro2016towards,zhang2016single,sam2017switching,sindagi2017generating,boominathan2016crowdnet} have demonstrated significant improvements over the traditional methods. Interested readers are referred to \cite{sindagi2017survey} for a comprehensive survey of existing methods. Most of the CNN-based methods   \cite{zhang2016single,onoro2016towards} addressed the issue of  scale variation using different architectures.   Zhang \etal \cite{zhang2016single} used multi-column architecture, with each column having different receptive field sizes. 
Sam \etal \cite{sam2017switching} built upon the multi-column networks, where they trained a Switching-CNN network to automatically choose the most optimal regressor among several independent regressors for a particular input patch.  Sindagi \etal \cite{sindagi2017generating} proposed Contextual Pyramid CNN (CP-CNN), where they demonstrated significant improvements by fusing local and global context through classification networks. 
Babu \etal \cite{babu2018divide} proposed a mechanism which involved  automatically growing CNN to incrementally increase the network capacity. In another interesting approach, Liu \etal \cite{liu2018leveraging} proposed to leverage unlabeled data for counting by introducing a learning to rank framework.  Recently, Li \etal \cite{li2018csrnet} proposed CSR-Net, that consists   of two  components: a front end CNN-based feature extractor and  a dilated CNN for the back-end.  

\begin{figure*}[t!]
	\begin{center}
		\vskip-20pt
		\includegraphics[width=1\linewidth]{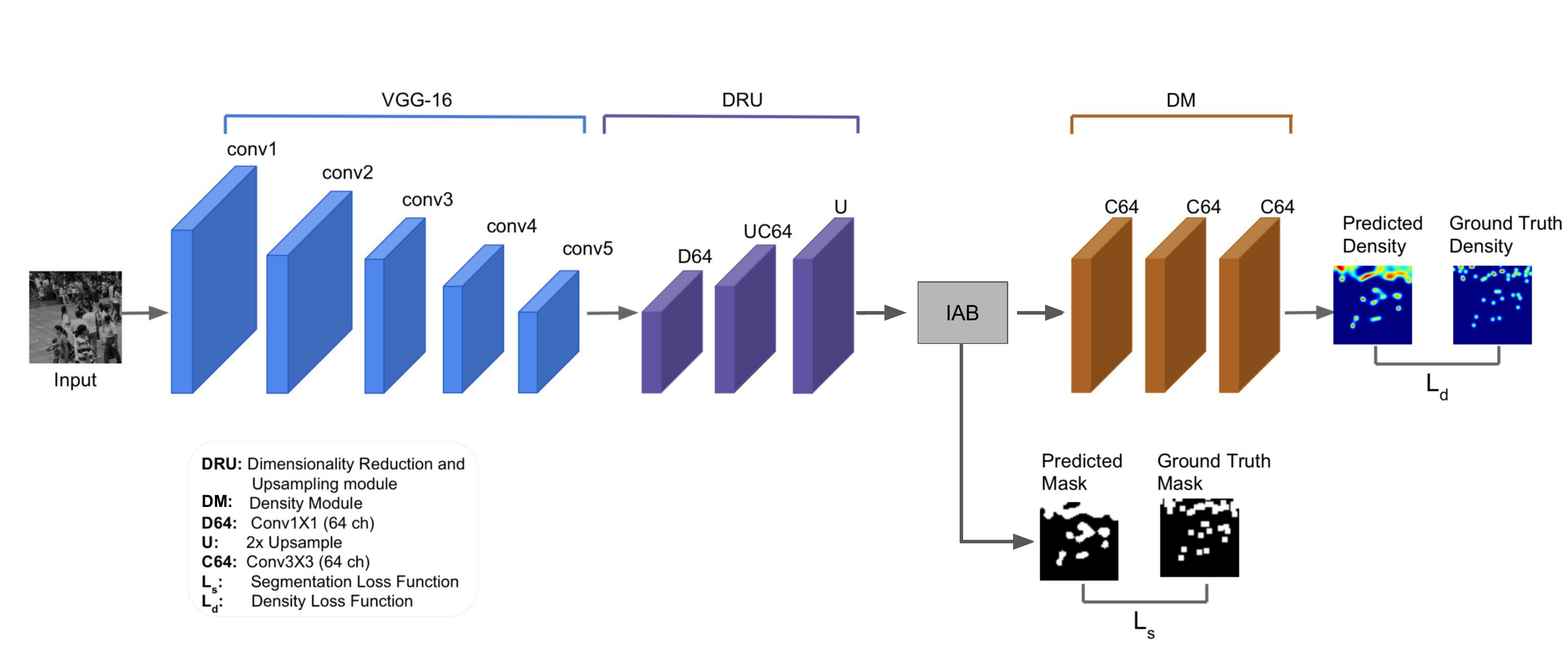}
	\end{center}
	\vskip -25pt \caption{Overview of the proposed Inverse Attention Guided  Deep Crowd Counting Network (IA-DCCN).}
	\label{fig:arch}
\end{figure*}

\section{Inverse Attention Guided  Deep Crowd Counting Network (IA-DCCN)}
\label{sec:method}

Fig. \ref{fig:arch} provides an overview of the proposed Inverse Attention Guided Deep Crowd Counting Network (IA-DCCN) which is based on the VGG-16 network. We include an inverse attention block (IAB)  with the objective of enriching the feature maps from VGG-16, thereby resulting in substantial improvements in the performance. This inverse attention block  aims to encode segmentation masks into the feature maps due to which the counting task becomes considerably easier. Additionally, we employ a simple hard-mining technique, that effectively samples the training data due to which appreciable gains are observed. Details of the proposed method and its various components are described in the following sub-sections.

\subsection{Base network}
\label{ssec:basenetwork}
As illustrated in Fig. \ref{fig:arch}, the base network consists of three parts: (i) first five convolutional blocks (conv1-conv5) from the VGG-16 architecture, (ii) dimensionality reduction and upsampling (DRU) module that reduces the dimensionality of feature maps from VGG-16 along the depth to 64 channels and upsamples them, and (iii) density module (DM) a set of three conv layers (with 64 channels and 3$\times$3 filters) to perform the density estimation. Note that the network is fully convolutional and hence, it can be used on images of any size. The entire network regresses on the input image to produce a density map which indicates the number of people per pixel. This density map, when summed over all the pixels, provides  an estimate of the number of people in the input image. The conv layers belonging to VGG-16 architecture are initialized with pre-trained weights, where as the conv layers in the density estimation network are randomly initialized with $\mu=0$ and $std=0.01$. The network is trained by minimizing the following loss function:
\begin{equation}
\label{eq:lossdensity}
L_d = \frac{1}{N}\sum_{i=1}^{N}\|F_d(X_i,\Theta) - D_{i}\|_2,
\end{equation}
where, $N$ is number of training samples, $X_i$ is the $i$\textsuperscript{th} input image, $F_d(X_i, \Theta)$ is the estimated density,  $D_i$ is the $i$\textsuperscript{th} ground-truth density and it is calculated by summing a 2D Gaussian kernel centered at every person's location $x_g$ as follows: 
\begin{equation}
\label{eq:densitymap}
D_i(x) = \sum_{{x_g \in S}}\mathcal{N}(x-x_g,\sigma),
\end{equation}
where $\sigma$ is scale parameter of 2D Gaussian kernel and $S$ is the set of all points at which people are located.  Following \cite{zhang2016single}, the density map generated by the network is $1$/$4^{th}$ of the input image resolution.

\subsection{Segmentation infusion via Inverse Attention}
\label{ssec:segmentation}

The base-network, although very simple, achieves significantly low count errors and the results are better/comparable with respect to recent state-of-the-art methods \cite{sam2017switching,sindagi2017generating}. In order to further boost the performance, we propose to incorporate segmentation information into the counting network. A naive idea would be add to segmentation loss layer after an intermediate block in the network and train the network in a multi-task fashion. This would be  similar to recent works like \cite{ranjan2016hyperface,hariharan2014simultaneous,hariharan2014simultaneous,dai2016instance} that learned different tasks simultaneously.

While this method results in a better performance as compared to the base network, we propose a more sophisticated method that uses inverse attention to  incorporate segmentation  information. For this,  we draw inspiration from the recent work on tasks like image captioning, super-resolution, classification, visual question answering \cite{chen2017sca,song2018mask} that use different forms of attention mechanisms to learn the features more effectively. Specifically, we introduce an inverse attention block (IAB) on top of the DRU module in the counting network as shown in Fig. \ref{fig:arch}.  

\begin{figure}[t!]
	\begin{center}
		\includegraphics[width=1\linewidth]{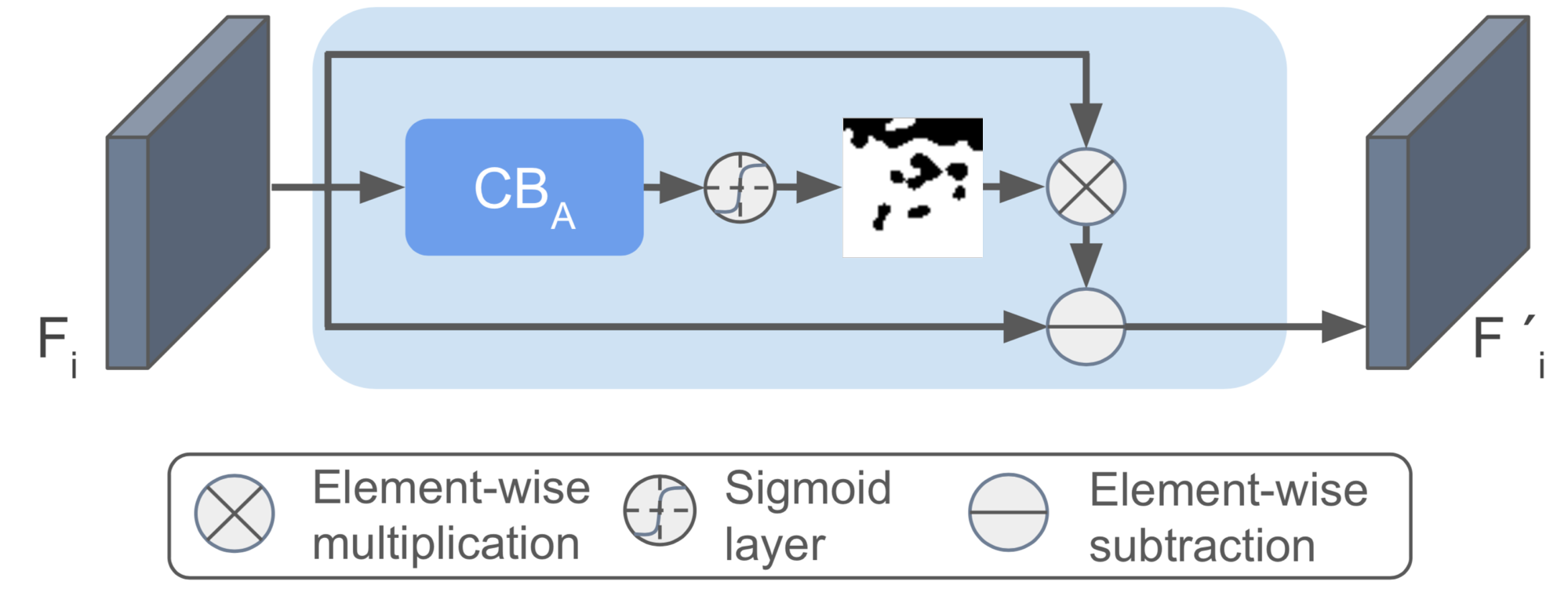}
	\end{center}
	\vskip -18pt \caption{Inverse attention block.}
	\label{fig:iab}
\end{figure}

Fig \ref{fig:iab} illustrates the mechanism of the inverse attention block. Specifically,  the $IAB$ takes feature maps  ($F$)  from the DRU as input and forwards them through a conv block $CB_A$ to  estimate background regions  (which we call as inverse attention map - $A^{-1}$) in the input image. $CB_A$ is defined by  \textit\{conv\sub{512,32,1}-relu-conv\sub{32,32,3}-relu-conv\sub{32,1,3}\}\footnote{ \label{fn:conv}\textit{conv\sub{N\sub{i},N\sub{o},k}} denotes conv layer (with \textit{N\sub{i}} input channels, \textit{N\sub{o}} output channels, \textit{k}$\times$\textit{k} filter size), \textit{relu} denotes ReLU activation}. Feature maps $F$ weighted by the inverse attention map are then subtracted from $F$  to suppress the background regions \ie, $$F' = F -  F \odot A^{-1},$$
where $F'$ is the attended feature map which is then forwarded through the density map module.

While the existing attention-based work learn the attention maps in a self-supervised manner, we instead use the ground-truth density maps to generate ground-truth inverse attention maps for supervising the inverse attention block. To generate the ground-truth, the density maps are thresholded and inverted. Note that by learning to estimate the background regions we are automatically suppressing the background information  from the feature maps of the DRU, hence, making it easier for the density module (DM) to learn the features  more effectively.  Fig. \ref{fig:fmap} illustrates the feature maps before and after enrichment using $IAB$. It can be clearly observed that the use of inverse attention block aids in better feature learning.



The entire network is trained in a multi-task fashion by simultaneously minimizing the density loss and the segmentation loss. Formally, the overall loss ($L$) is defined as follows:$L = L_d + \lambda_sL_s$
\begin{equation}
\label{eq:finalloss}
L = L_d + \lambda_sL_s,
\end{equation}

where, $L_d$ is the density loss (Eq. \ref{eq:lossdensity}), $L_s$ is segmentation loss that is used for training the features of the inverse attention block, and $\lambda_s$ is weighting factor for the segmentation loss. $L_s$ is pixel-wise cross entropy error between estimated mask and ground-truth mask. Note that, the method does not require any additional labeling and it uses weakly annotated head/person regions based on the existing labels to compute the segmentation loss. 
The ground-truth mask is generated by thresholding the ground-truth density map. Bascially, the pixels that contain head regions are labeled as 1 (foreground), and otherwise as 0 (background). In spite of these annotations being noisy, the use of segmentation information results in considerable gains.

\subsection{Hard sample mining (HSM)}
\label{ssec:hsm}
Recent methods such as \cite{shrivastava2016training,loshchilov2015online}  have demonstrated that effective sampling of data by selecting harder samples improves the classification performance of the network. Similar to these work, we employ an offline hard mining technique to train the network. This process, used to select samples from the training set every 5 epochs, involves the following steps: (i) compute the histogram of error on the entire training data, (ii) find the mode ($T$) of this error distribution (iii) training samples with  $error > T$ are considered as hard samples and selected for training. This sample selection technique is effective in lowering the count error by an appreciable margin.

\section{Experiments and results}
\label{sec:experiments}

In this section, we first describe the training and implementation specifics followed by a detailed ablation study to understand the effects of different components in the proposed network. We chose the ShanghaiTech dataset \cite{zhang2016single} to perform the ablation study as it contains significant variations in count and scale. Finally, we compare the results of the proposed method against several recent approaches on three publicly available datasets containing congested scenes. (ShanghaiTech, \UCF\; \cite{idrees2013multi}, UCF-QNRF \cite{idrees2018composition}).

\subsection{Training and implementation details} 

\noindent The network is trained end-to-end using the Adam optimizer with a learning rate of 0.00005 and a momentum of 0.9 on a single NVIDIA GPU Titan Xp. 10 \% of the training set is set aside for validation purpose. The final training dataset is formed by cropping patches of size 224$\times$224 from 9 random locations from each image. Furthermore, data augmentation is performed by randomly flipping the images (horizontally) and adding random noise. Since the network is fully convolutional, image of any arbitrary size or resolution can be input to the network. Similar to earlier work, the count performance is measured using the following metrics:
\begin{equation}
\label{eq:count_error}
\nonumber  MAE = \frac{1}{N}\sum_{i=1}^{N}|y_i-y'_i|, \;\;
MSE = \sqrt{\frac{1}{N}\sum_{i=1}^{N}|y_i-y'_i|^2},
\end{equation}
where, $MAE$ is mean absolute error, $MSE$ is mean squared error, $N$ is the number of test samples, $y_i$ is the ground-truth count and $y'_i$ is the estimated count corresponding to the $i^{th}$ sample. \newline

%
%
%
%
%
%
\begin{table*}[ht!]
	\centering
	\caption{Results of the ablation study on the ShanghaiTech Part A and Part B datasets. Figures in braces indicate the percentage improvement in error over previous configuration.}
	\vskip-8pt
	\label{tab:ablation}
	\resizebox{.65\linewidth}{!}{
		\begin{tabular}{|l|c|c|c|c|}
			\hline
			& \multicolumn{2}{c|}{Part A}  & \multicolumn{2}{c}{Part B}  \\ \hline
			Configuration                    & MAE          & MSE           & MAE          & MSE           \\\hline
			Base network                    & 76.7                & 119.1                & 17.3                 & 22.5         \\
			Base network+S               & 71.2 (7.1\%) & 117.5 (1.3\%)  & 15.0 (13.3\%) & 21.0 (6.7\%) \\
			Base network+IAB            & 68.1 (4.3\%) & 114.5 (2.5\%)  & 13.6 (9.3\%) & 19.6 (6.7\%) \\
			Base network+IAB+HSM & 66.9 (1.8\%) & 108.5 (5.2\%) & 10.2 (25.0\%) & 16.0 (18.3\%)  \\ \hline
		\end{tabular}
	}
\end{table*}

\subsection{Architecture ablation} 

\noindent To understand the effectiveness of the various modules present in the network, we perform experiments with the  different settings using the ShanghaiTech dataset (Part A and Part B). This dataset consists of 2 parts with Part A containing 482 images and Part B containing 716 images and a total of 330,165 head annotations. Both parts have training and test subsets.  Due to various challenges such as high density crowds, large variations in scales, presence of occlusion, etc, we chose to perform the ablation study on this dataset. 

The results of these experiments are tabulated in Table \ref{tab:ablation}. It  can be observed that the base network, consisting of VGG-16 conv layers along with DRU module and density module (described in Section \ref{ssec:basenetwork}), does not provide the optimal performance. With the addition of  the segmentation loss layer (Base network + S), we can observe    an improvement of $\sim$7.1\%/1.3\% in MAE/MSE on Part-A and $\sim$15.0\%/6.7\% in MAE/MSE on Part-B over the base network. While this naive method of infusing segmentation network results in considerable improvements in the error, we show that there is still room for further improvements with the experiment where we incorporated the inverse attention block after the DRU module (Base network + IAB). The $IAB$ module results in an improvement of $\sim$4.3\%/2.5\% in MAE/MSE on Part-A and $\sim$9.3\%/6.7\% in MAE/MSE on Part-B over the naive method.

Fig. \ref{fig:fmap} visualizes the feature maps from the DRU module  network with/without feature enrichment via segmentation infusion using inverse attention. Effects of incorporating segmentation information into the counting network can be clearly observed. This considerable reduction in the count error confirms our intuition that  segmentation guided inverse  can be used to aid the counting  task, by introducing high-level foreground background knowledge into the feature maps of the VGG-16 network. $\lambda_s$ in Eq. \ref{eq:finalloss} is set equal to 0.1 based on cross-validation.

Finally, we trained the entire network with hard sample mining (Base network + IAB + HSM as described in Section \ref{ssec:hsm}), where samples for training were selected based on the error at every fifth epoch. For this configuration, we observed an improvement of $\sim$1.8\%/5.2\% in MAE/MSE on Part A and $\sim$25.0\%/18.3\% in MAE/MSE on Part B over the base network with inverse attention.   Hence, it can be concluded that hard sample mining is effective and provides appreciable gains on both parts of the dataset.

\begin{table}[t!] 
	\centering 		
	\caption{Comparison of results on the ShanghaiTech dataset.} 
	\vskip-8pt
	\label{tab:shtech}
	\resizebox{.951\linewidth}{!}{
		\begin{tabular}{|l|c|c|c|c|}
			\hline
			& \multicolumn{2}{c|}{Part A} & \multicolumn{2}{c}{Part B} \\ \hline
			Method          & MAE          & MSE          & MAE          & MSE          \\ \hline
			Cascaded-MTL \cite{sindagi2017cnnbased}  (AVSS '17)         & 101.3        & 152.4        & 20.0         & 31.1         \\ 
			Switching-CNN \cite{sam2017switching}   (CVPR '17)        & 90.4        & 135.0        & 21.6         & 33.4         \\ 
			TopDownFeedack \cite{sam2018top}   (AAAI '18)        & 97.5        & 145.1       & 20.7         & 32.8         \\ 
			CP-CNN \cite{sindagi2017generating} (ICCV '17)          & 73.6        & \textbf{106.4}       & 20.1         & 30.1         \\ 
			IG-CNN \cite{babu2018divide}   (CVPR '18) & 72.5        & 118.2        & 13.6         & 21.1         \\ 
			CSR-Net \cite{li2018csrnet}   (CVPR '18) & 68.1        & 115.0        & 10.6         & 16.0         \\ 
			IA-DCCN (ours) & \textbf{66.9}        & {108.4}        & \textbf{10.2}         & \textbf{16.0}         \\ \hline
		\end{tabular}
	}
	
\end{table}

\begin{figure*}[t!]	
	\centering	
	\includegraphics[width=0.3\linewidth]{figures/sa/IMG_2}
	\includegraphics[width=0.3\linewidth]{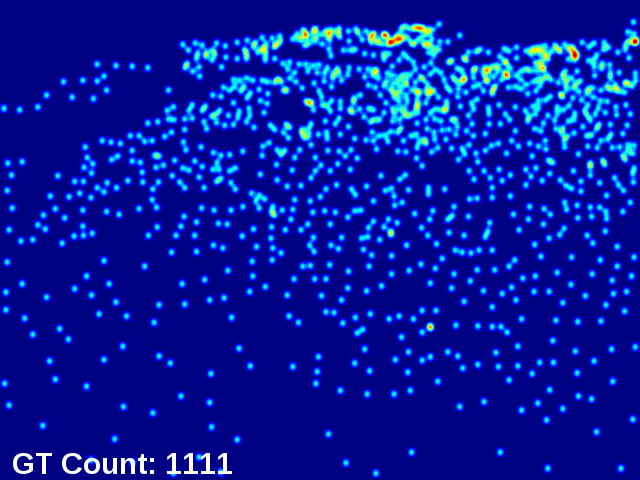}
	\includegraphics[width=0.3\linewidth]{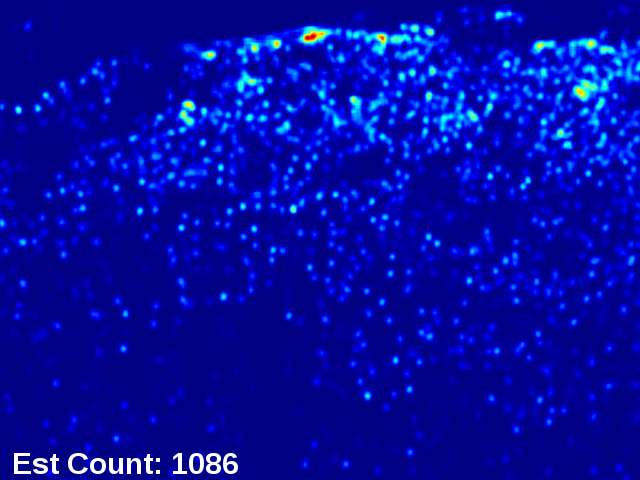}\\
	\includegraphics[width=0.3\linewidth]{figures/sa/IMG_74}
	\includegraphics[width=0.3\linewidth]{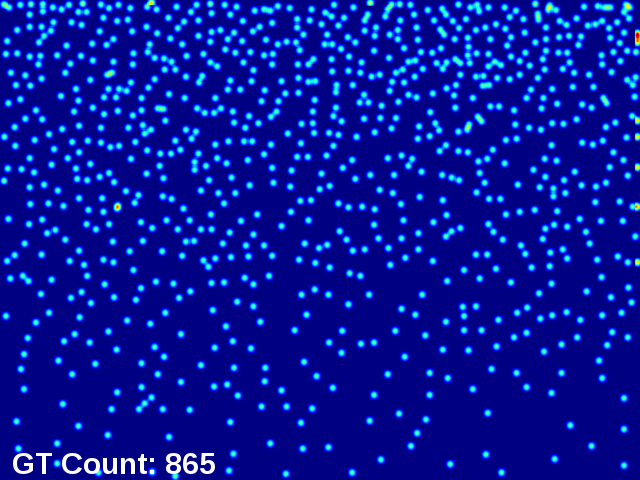}
	\includegraphics[width=0.3\linewidth]{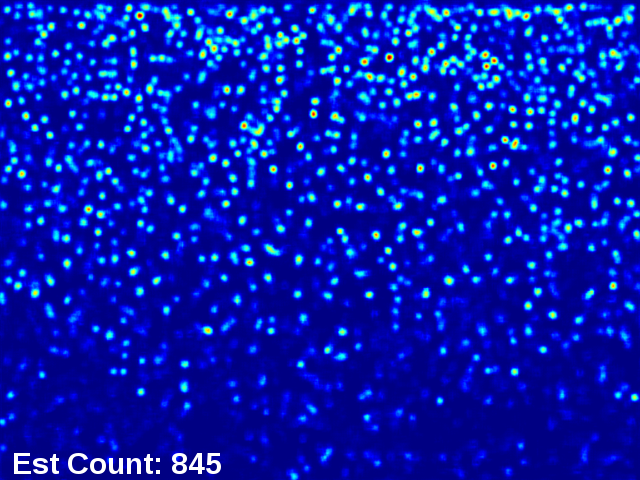}	\\
	(a) \hskip120pt(b)  \hskip120pt(c)
	\vskip -8pt\caption{Sample results of the proposed method on the  ShanghaiTech dataset \cite{zhang2016single}. \textit{(a)} Input. \textit{(b)} Ground truth \textit{(c)} Estimated density map.}
	\label{fig:shtechresults1}	
\end{figure*} 	

\begin{figure*}[t!]	
	
	\includegraphics[width=0.3\linewidth]{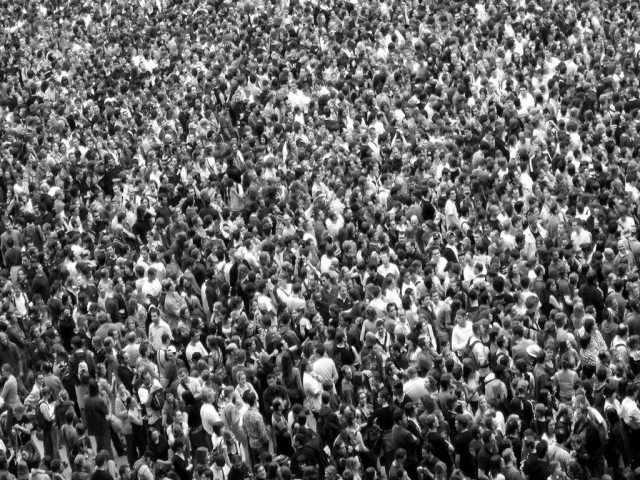}
	\includegraphics[width=0.3\linewidth]{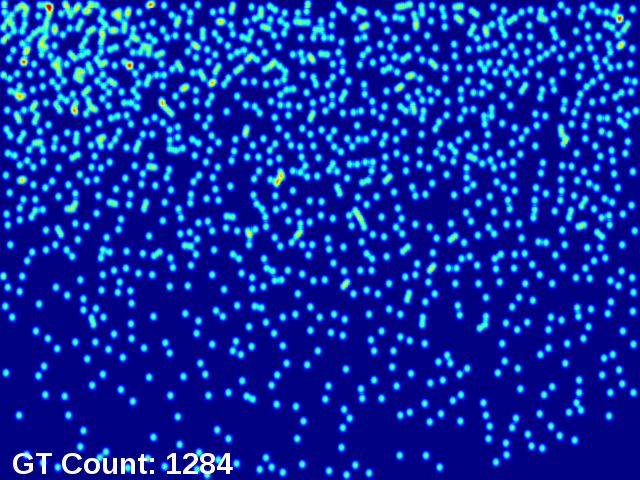}
	\includegraphics[width=0.3\linewidth]{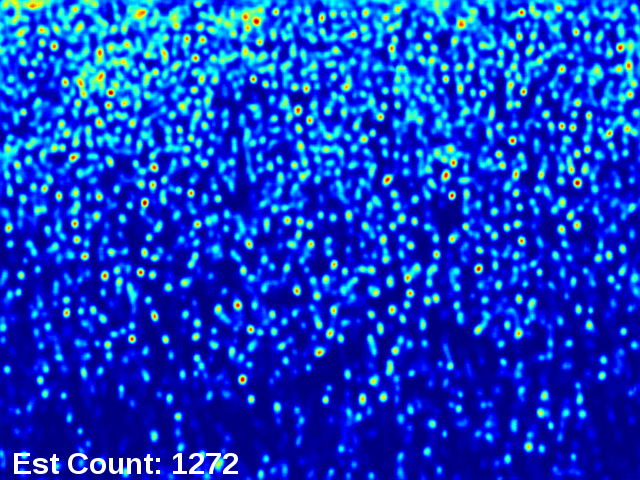}
	
	\includegraphics[width=0.3\linewidth]{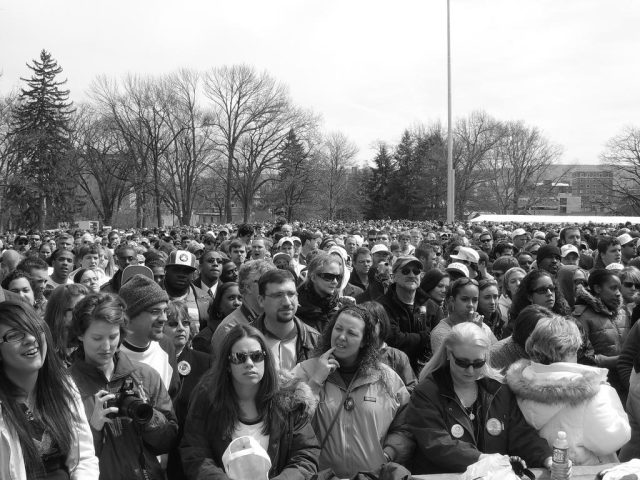}
	\includegraphics[width=0.3\linewidth]{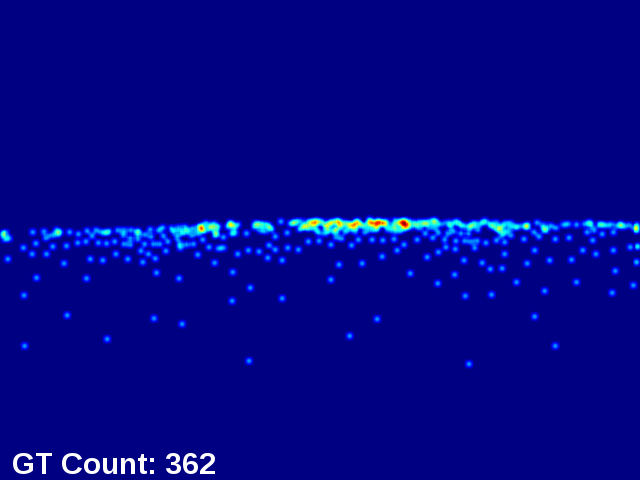}
	\includegraphics[width=0.3\linewidth]{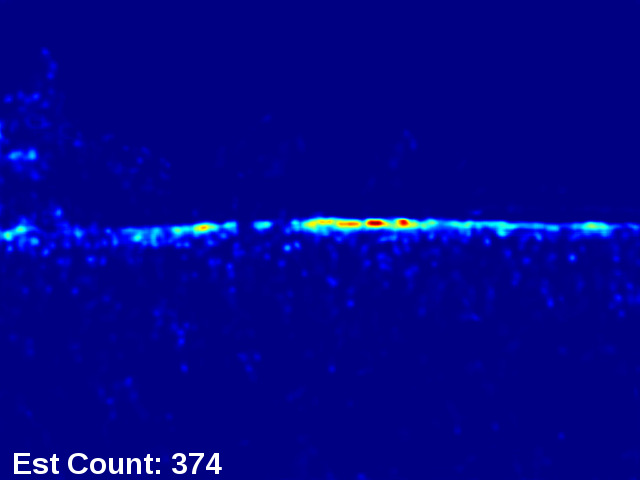}
	\\
	(a) \hskip120pt(b)  \hskip120pt(c)
	\vskip -8pt\caption{Sample results of the proposed method on the \UCF\; dataset \cite{idrees2013multi}. \textit{(a)} Input. \textit{(b)} Ground truth \textit{(c)} Estimated density map.}
	\label{fig:shtechresults2}	
\end{figure*}

\subsection{Comparison with recent methods} 

\noindent For comparison with various methods on different datasets, the entire network (IA-DCCN) is trained with hard sample mining.

\noindent\textbf{ShanghaiTech.} The proposed method is compared with four recent approaches ( Cascaded-MTL \cite{sindagi2017cnnbased}, Switching-CNN \cite{sam2017switching}, CP-CNN \cite{sindagi2017generating}, Top-down feedback \cite{sam2018top}, IG-CNN \cite{babu2018divide} and CSR-Net \cite{li2018csrnet}) on Part A and Part B of the ShanghaiTech dataset and the results are presented in Table \ref{tab:shtech}. The proposed IA-DCCN method achieves the lowest error rate in terms of MAE/MSE as compared to all recent methods on both parts of the dataset.  
Sample density estimation results are shown in Fig. \ref{fig:shtechresults1}. From these results, it can be noted that the proposed method is able to achieve encouraging results while being simple to train as compared to existing approaches.

\begin{table}[ht!]
	\centering
	\caption{Comparison of results on the \UCF\; dataset.}
	\label{tab:resultsucf}
	\resizebox{0.75\linewidth}{!}{
		\begin{tabular}{|l|c|c|}
			\hline
			Method & MAE & MSE \\			\hline
			Cascaded-MTL \cite{sindagi2017cnnbased} (AVSS '17) & 322.8 & 397.9 \\		
			Switching-CNN \cite{sam2017switching} (CVPR '17) & 318.1 & 439.2 \\
			CP-CNN \cite{sindagi2017generating} (ICCV '17) & 295.8 & \textbf{320.9} \\
			TopDownFeedback \cite {sam2018top} (AAAI '18) & 354.7 & 425.3 \\
			IG-CNN \cite{babu2018divide} (CVPR '18) & 291.4 & 349.4 \\
			CSR-Net \cite{li2018csrnet} (CVPR '18) & 266.1 & 397.5 \\
			IA-DCCN (ours) & \textbf{264.2} & 394.4 \\
			\hline
		\end{tabular}
	}
\end{table}

\begin{table}[ht!]
	\centering
	\caption{Comparison of results on the UCF-QNRF datastet.}
	\label{tab:resultsucf}
	\resizebox{0.7\linewidth}{!}{
		\begin{tabular}{|l|c|c|}
			\hline
			Method & MAE & MSE \\			\hline
			Idrees \etal \cite{idrees2013multi} (CVPR '13) & 315.0 & 508.0 \\
			Zhang \etal \cite{zhang2015cross} (CVPR '15) & 277.0 & 426.0 \\
			Cascaded-MTL  \cite{sindagi2017cnnbased} (AVSS '17) & 252.0 & 514.0 \\		
			Switching-CNN \cite{sam2017switching} (CVPR '17) & 228.0 & 445.0 \\
			Idrees \etal \cite{idrees2018composition} (ECCV '18) & {132.0} & {{191.0}} \\
			IA-DCCN (ours) & {\textbf{125.3}} & {\bf{185.7 }}\\
			\hline
		\end{tabular}
	}
\end{table}

\noindent\textbf{\UCF.}  The UCF\textunderscore CC\textunderscore 50 dataset \cite{idrees2013multi} is a relatively smaller dataset with   50 annotated images of different resolutions and aspect ratios. We used the standard 5-fold cross-validation protocol discussed in \cite{idrees2013multi} to  evaluate  the proposed method. Results are compared with several recent approaches: Cascaded-MTL \cite{sindagi2017cnnbased}, Switching-CNN \cite{sam2017switching}, CP-CNN \cite{sindagi2017generating}, Top-down feedback \cite{sam2018top}, IG-CNN \cite{babu2018divide} and  CSR-Net \cite{li2018csrnet}.  The results are tabulated in Table \ref{tab:resultsucf}.
It can be observed that the proposed method achieves the lowest MAE error as compared to the recent methods. Although the proposed approach performs slightly worse in terms of MSE as compared to CP-CNN \cite{sindagi2017generating}, it is important to note that the MSE error is  comparable to the existing approaches.  Additionally, we believe that these results are especially significant considering the simplicity of the proposed approach. Sample density estimation results are shown in Fig. \ref{fig:shtechresults2}.

\noindent\textbf{UCF-QNRF}: The UCF-QNRF \cite{idrees2018composition} is a recent  dataset that contains around 1200 images with approximately 1.2 million annotations. The results of the proposed method  on this dataset as compared with recent methods (\cite{idrees2013multi},\cite{zhang2016single},\cite{sindagi2017cnnbased}) are shown in Table \ref{tab:resultsucf}. The proposed method is compared against five different approaches: \cite{idrees2013multi},   \cite{zhang2016single},   \cite{sindagi2017cnnbased},\cite{sam2017switching}, and \cite{idrees2018composition}. It can be observed that the proposed method outperforms other methods by a considerable margin.   

\subsubsection{Inference speed} 

\noindent To evaluate the inference speed of the proposed approach, we run IA-DCCN on our machine which is equipped with Intel Xeon E5-2620v4@2.10GHz and an NVIDIA Titan Xp GPU. The run times are reported in Table \ref{tab:speed} for different resolutions ranging from 320$\times$240 to 1600$\times$1200. It can be observed that the proposed method is  efficient and is able to run at $\sim$76 fps while processing high resolution images (1600$\times$1200). Note that the majority of processing time is taken up by the VGG-16 network. 

\begin{table}[h!]
	\centering
	\caption{Inference time for different resolutions in msec.}
	\label{tab:speed}
	\resizebox{1\linewidth}{!}{
		\begin{tabular}{|l|c|c|c|c|}
			\hline
			Res (W$\times$H) & 320$\times$240 & 640$\times$480 & 1280$\times$960 & 1600$\times$1200 \\ \hline
			IA-DCCN (ours)               & 2.7            & 4.8            & 8.9             & 13.1             \\ \hline
		\end{tabular}
	}
\end{table}
\vspace{-0.25in} 

\section{Conclusions}
\label{sec:conclusion}

We presented a very simple, yet effective crowd counting approach based on the VGG-16 network and inverse attention, referred to as Inverse Attention Guided Deep Crowd Counting Network (IA-DCCN). The proposed approach aims to infuse segmentation information into the counting network via  an inverse attention mechanism. This infusion of segmentation maps into the network enriches the feature maps of VGG-16 network due to which the background information in the feature maps get suppressed, making the counting task rather easier. In contrast to existing approaches that employ complex training process, the proposed approach is a single-stage training framework and achieves significant improvements over the recent methods while being computationally fast.  

\section*{Acknowledgments}
This work was supported by US Office of Naval Research (ONR) Grant YIP N00014-16-1-3134.

{\small
	\bibliographystyle{ieee}
	\bibliography{egbib}
}

\end{document}